\documentclass[a4paper,11pt]{article}
\pdfoutput=1 
\usepackage{graphicx}  
\usepackage{url}       
\usepackage{times}
\usepackage{natbib}
\usepackage[margin=25mm]{geometry}

\usepackage{latexsym}
\usepackage{amsmath}
\usepackage{amssymb} 
\usepackage{multirow}
\usepackage{paralist}

\usepackage{subfigure}

\newcommand{\semantics}[1]{[\![ #1 ]\!]}
\newcommand{\newcite}{\citet}
\newcommand{\shortcite}{\citet}

\usepackage{authblk}

\title{Multi-Step Regression Learning for Compositional Distributional Semantics}
\date{}

\author[$\ast$]{E.~Grefenstette}
\author[$\dag$]{G.~Dinu}
\author[$\ddag$]{Y.~Zhang}
\author[$\ast$]{M.~Sadrzadeh}
\author[$\dag$]{M.~Baroni}

\affil[$\ast$]{University of Oxford Department of Computer Science}
\affil[$\dag$]{University of Trento Center for Mind/Brain Sciences}
\affil[$\ddag$]{University of Tokyo Department of Computer Science}

\begin{document}
\maketitle
\thispagestyle{empty}
\pagestyle{empty}
\begin{abstract}
  We present a model for compositional distributional semantics
  related to the framework of~\citet{Coecke:etal:2010}, and emulating
  formal semantics by representing functions as tensors and arguments
  as vectors. We introduce a new learning method for tensors,
  generalising the approach of~\citet{Baroni:Zamparelli:2010}. We
  evaluate it on two benchmark data sets, and find it to outperform
  existing leading methods. We argue in our analysis that the nature
  of this learning method also renders it suitable for solving more subtle
  problems compositional distributional models might face.
\end{abstract}

\section{Introduction}
\label{sec:introduction}

The staggering amount of machine readable text available on today's Internet calls for increasingly powerful text and language processing methods. This need has fuelled the search for more subtle and sophisticated representations of language meaning, and methods for learning such models. Two well-researched but \emph{prima-facie} orthogonal approaches to this problem are formal semantic models and distributional semantic models, each complementary to the other in its strengths and weaknesses.

Formal semantic models generally implement the view of \newcite{Frege:1892}---that the semantic content of an expression is its logical form---by defining a systematic passage from syntactic rules to the composition of parts of logical expressions. This allows us to derive the logical form a of sentence from its syntactic structure \citep{Montague1970EFL}. These models are fully compositional, whereby the meaning of a phrase is a function of the meaning of its parts; however, as they reduce meaning to logical form, they are not necessarily adapted to all language processing applications such as paraphrase detection, classification, or search, where topical and pragmatic relations may be more relevant to the task than equivalence of logical form or truth value. Furthermore, reducing meaning to logical form presupposes the provision of a logical model and domain in order for the semantic value of expressions to be determined, rendering such models essentially \emph{a priori}.

In contrast, distributional semantic models, suggested by
\newcite{firth1957papers}, implement the linguistic philosophy of
\newcite{Wittgenstein:1953} stating that meaning is associated with
use, and therefore meaning can be learned through the observation of
linguistic practises. In practical terms, such models learn the
meaning of words by examining the contexts of their occurrences in a
corpus, where `context' is generally taken to mean the tokens with
which words co-occur within a sentence or frame of $n$ tokens. Such
models have been successfully applied to various tasks such as
thesaurus extraction \citep{Grefenstette:1994} and essay grading
\citep{Landauer:Dumais:1997,Dumais:2003}. However, unlike their formal
semantics counterparts, distributional models have no explicit
canonical composition operation, and provide no way to integrate
syntactic information into word meaning combination to produce
sentence meanings.

In this paper, we present a new approach to the development of
\emph{compositional} distributional semantic models, based on earlier
work by~\newcite{Baroni:Zamparelli:2010},~\newcite{Coecke:etal:2010}
and~\newcite{Grefenstetteetal2011}, combining features from the
compositional distributional framework of the latter two with the
learning methods of the former. In Section \ref{sec:related-work} we
outline a brief history of approaches to compositional distributional
semantics. In Section \ref{sec:model} we overview a tensor-based
compositional distributional model resembling traditional formal
semantic models. In Section \ref{sec:learning} we present a new
multi-step regression algorithm for learning the tensors in this
model. Sections \ref{sec:experiments}--\ref{sec:exp-intransitives}
present the experimental setup and results of two experiments
evaluating our model against other known approaches to
compositionality in distributional semantics, followed by an analysis
of these results in Section \ref{sec:discussion}. We conclude in
Section \ref{sec:conclusion} by suggesting future work building on the
success of the model presented in this paper.

\section{Related work}
\label{sec:related-work}

Although researchers tried to derive sentence meanings by composing
vectors since the very inception of distributional semantics, this
challenge has attracted special attention in recent
years. \shortcite{Mitchell:Lapata:2008,Mitchell:Lapata:2010} proposed
two broad classes of composition models (additive and multiplicative)
that encompass most earlier and related proposals as special
cases. The simple additive method (summing the vectors of the words in
the sentence or phrase) and simple multiplicative method
(component-wise multiplication of the vectors) are straightforward and
empirically effective instantiations of the general models. We
re-implemented them here as our Add and Multiply methods (see Section
\ref{sec:composition-methods} below).

In formal semantics, composition has always been modeled in terms of
function application, treating certain words as functions that operate
on other words to construct meaning incrementally according to a
calculus of composition that reflects the syntactic structure of
sentences
\citep{Frege:1892,Montague1970EFL,Partee:2004}. \newcite{Coecke:etal:2010}
have proposed a general formalism for composition in distributional
semantics that captures the same notion of function
application. Empirical implementations of Coecke's et al.'s 
formalism have been developed by~\newcite{Grefenstetteetal2011} and
tested by
\shortcite{Grefenstette:Sadrzadeh:2011a,Grefenstette:Sadrzadeh:2011b}. In
the methods they derive, a verb with $r$ arguments is a rank $r$
tensor to be combined via component-wise multiplication with the
Kronecker product of the vectors representing its arguments, to obtain
another rank $r$ tensor representing the sentence:
\[ S = V \odot (\mathbf{a}_1 \otimes \mathbf{a}_2 \otimes ... \otimes
\mathbf{a}_r) \] \newcite{Grefenstette:Sadrzadeh:2011b} propose
various ways to estimate the components of verb tensors in the
two-argument (transitive) case, with the simple method of constructing
the rank 2 tensor (matrix) by the Kronecker product of a corpus-based
verb vector with itself giving the best results. The Kronecker method
outperformed the best method of
\newcite{Grefenstette:Sadrzadeh:2011a}, referred to as the Categorical
model.  We re-implement the Kronecker method for our experiments
below. It was not possible to efficiently implement the Categorial
method across our large corpus, but we still provide a meaningful
indirect comparison with this method.



\newcite{Baroni:Zamparelli:2010} propose a different approach to
function application in distributional space, that they apply to
adjective-noun composition (see also \newcite{Guevara:2010} for
similar ideas). Adjectives are functions, encoded as linear maps, that
take a noun vector as input and return another nominal vector
representing the composite meaning as output. In linear algebraic terms,
adjectives are matrices, and composition is matrix-by-vector
multiplication:
\[ \mathbf{c} = A \times \mathbf{n} \]
\newcite{Baroni:Zamparelli:2010} estimate the adjective matrices by
linear regressions on corpus-extracted examples of their input and
output vectors. In this paper, we derive their approach as a special
case of a more general framework and extend it, both theoretically
and empirically, to two-argument functions (transitive verbs), as well
as testing the original single argument variant in the verbal
domain. Our generalisation of their approach is called Regression in
the experiments below.

In the MV-RNN model of \cite{socherEMNLP12}, \emph{all} words and
phrases are represented by both a vector and a matrix, and composition
also involves a non-linear transformation. When two expressions are
combined, the resulting composed vector is a non-linear function of
the concatenation of two linear transformations (multiplying the first
element matrix by the second element vector, and \emph{vice versa}).
In parallel, the components of the matrices associated with the
resulting phrase are linear combinations of the components of the
input matrices. 
Socher and colleagues show that MV-RNN reaches state-of-the-art
performance on a variety of empirical tasks.

While the proposal of Socher et al.~is similar to our approach in many
respects, including syntax-sensitivity and the use of matrices in the
calculus of composition, there are three key differences. The first is
that MV-RNN requires task-specific labeled examples to be trained for
each target semantic task, which our framework does not, attempting to
achieve greater generality while relying less on manual
annotation. 
The second difference, more theoretical in nature, is that all
composition in MV-RNN is pairwise, whereas we will present a model of
composition permitting functions of larger arity, allowing the
semantic representation of functions that take two or more arguments
simultaneously. Finally, we follow formal semantics in treating
certain words as functions and other as arguments (and can thus
directly import intuitions about the calculus of composition from
formal semantics into our framework), whereas Socher and colleagues
treat each word equally (as both a vector and a matrix). However, we
make no claim at this stage as to whether or not these differences can
lead to richer semantic models, leaving a direct comparison to future
work.

Several studies tackle word meaning in context, that is, how to adapt
the distributional representation of a word to the specific context in
which it appears
\citep[e.g.,][]{Dinu:Lapata:2010,Erk:Pado:2008,Thater:etal:2011}. We
see this as complementary rather than alternative to composition:
Distributional representations of single words should first be adapted
to context with these methods, and then composed to represent the
meaning of phrases and sentences.



\section{A general framework for distributional function application}
\label{sec:model}

A popular approach to compositionality in formal semantics is to
derive a formal representation of a phrase from its grammatical
structure by representing the semantics of words as functions and
arguments, and using the grammatical structure to dictate the order and
scope of function application. For example, formal semantic models in
the style of~\newcite{Montague1970EFL} will associate a semantic rule
to each syntactic rule in a context-free grammar. A sample formal
semantic model is shown here:
\begin{displaymath}
{
\footnotesize
\begin{tabular}{l|l}
\textbf{Syntax} & \textbf{Semantics}\\
\hline
S $\Rightarrow$ NP VP & $\semantics{S} \Rightarrow \semantics{VP} \left(\semantics{NP}\right)$\\
NP $\Rightarrow$ N & $\semantics{NP} \Rightarrow \semantics{N}$\\
N $\Rightarrow$ ADJ N & $\semantics{N} \Rightarrow \semantics{ADJ}\left(\semantics{N}\right)$\\
VP $\Rightarrow$ Vt NP & $\semantics{VP} \Rightarrow \semantics{Vt}\left(\semantics{NP}\right)$\\
VP $\Rightarrow$ Vi & $\semantics{VP} \Rightarrow \semantics{Vi}$\\
\end{tabular}
\qquad
\begin{tabular}{l|l}
\textbf{Syntax (cont'd)} & \textbf{Semantics (cont'd)}\\
\hline
Vt $\Rightarrow$ $\{\textit{verbs$_t$}\}$ & $\semantics{Vt} \Rightarrow \semantics{\textit{verb$_t$}} $\\
Vi $\Rightarrow$ $\{\textit{verbs$_i$}\}$ & $\semantics{Vi} \Rightarrow \semantics{\textit{verb$_i$}} $\\
ADJ $\Rightarrow$ $\{\textit{adjs}\}$ & $\semantics{ADJ} \Rightarrow \semantics{\textit{adj}}$\\
N $\Rightarrow$ $\{\textit{nouns}\}$ & $\semantics{N} \Rightarrow \semantics{\textit{noun}}$
\end{tabular}
}
\end{displaymath}
Following these rules, the parse of a simple sentence like `angry dogs
chase furry cats' yields the following interpretation:
$\semantics{\textit{chase}}( \semantics{\textit{furry}}
(\semantics{\textit{cats}})
)(\semantics{\textit{angry}}(\semantics{\textit{dogs}}))$.  This is a simple
model, where  typically lambda abstraction will be liberally
used to support quantifiers and argument inversion, but the key point
remains that the grammar dictates the translation from natural
language to the functional form, e.g.  predicates and logical relations. Whereas in formal semantics these functions have a set theoretic form, we
present here a way of defining them  as multilinear maps over
geometric objects. This geometric framework is also applicable to
other formal semantic models than that presented here. This is
particularly important, as the version of the model presented here is overly simple compared
to modern work in formal semantics (which, for example, apply NPs to
VPs instead of VPs to NPs, to model quantification), and only serves
as a model  frame within which we illustrate how our approach
functions.

The bijective correspondence between linear maps and matrices is a well known property in linear algebra: Every linear map $f: A \to B$ can be encoded as a $dim(B)$ by $dim(A)$ matrix $M$, and conversely every such matrix encodes a class of linear maps determined by the dimensionality of the domain and co-domain. The application of a linear map $f$ to a vector $\mathbf{v} \in A$ producing a vector $\mathbf{w} \in B$ is equivalent to the matrix multiplication:
\[
f(\mathbf{v}) = M \times \mathbf{v} = \mathbf{w}
\]
In the case of multilinear maps, this correspondence generalises to a correlation
between $n$-ary maps and rank $n+1$ tensors
\citep{bourbaki1989commutative,Lee:1997}. Tensors are 
 generalisations of  vectors and matrices; they have
\emph{larger degrees of freedom} referred to as tensor ranks, which is one for
vectors and two for matrices. To illustrate this generalisation,
consider how a row/column vector  may be written as the
weighted superposition (summation) of its basis elements: any vector
$\mathbf{v}$ in a
vector space $V$ with a fixed basis $\{\mathbf{b}_i\}_i$,  can be written
\[
\mathbf{v} = \sum_i{c^{\mathbf{v}}_i \mathbf{b}_i} = \left[c^{\mathbf{v}}_1,\, \ldots,\, c^{\mathbf{v}}_i,\, \ldots,\, c^{\mathbf{v}}_{dim(V)}\right]^{\top}
\]
Here,  the weights $c^{\mathbf{v}}_i$ are elements of the underlying
field (e.g.~$\mathbb{R}$), and thus vectors can be fully described by
such a one-index summation. Likewise, matrices, which are rank 2
tensors, can be seen as a collection of row vectors from some
space $V_r$ with basis $\{\mathbf{a}_i\}_i$, or of column vectors from some space
$V_c$ with basis $\{\mathbf{d}_j\}_j$. Such a matrix $M$ is an
element of the space $V_r \otimes V_c$, and can be fully described by
the two index summation:
\[
M = \sum_{ij}{c^{M}_{ij} \mathbf{a}_i \otimes \mathbf{d}_j}
\]
where, once again, $c^{M}_{ij}$ is an element of the underlying field
which in this case is simply the element from the $i$th row and
$j$th column of the matrix $M$, and the basis element $\mathbf{a}_i
\otimes \mathbf{d_j}$ of $V_r \otimes V_c$ is formed by a pair of
basis elements from $V_r$ and $V_c$. The number of indices (or degrees
of freedom) used to fully describe a tensor in this superposition
notation is its rank, e.g., a rank 3 tensor $T \in A \otimes B \otimes
C$ would be described by the superposition of weights $c^T_{ijk}$
associated with basis elements $\mathbf{e}_i \otimes \mathbf{f}_j
\otimes \mathbf{g}_k$.

The notion of matrix multiplication and inner product both generalise
to tensors as the non-com\-mutative tensor contraction operation
($\times$). For tensors $T \in A \otimes \ldots \otimes B \otimes C$
and $U \in C \otimes D \otimes \ldots \otimes E$, with bases
$\{\mathbf{a}_i \otimes \ldots \otimes \mathbf{b}_j \otimes
\mathbf{c}_k\}_{i \ldots jk}$ and $\{\mathbf{c}_k \otimes \mathbf{d}_l
\otimes \ldots \otimes \mathbf{e}_m\}_{kl \ldots m}$, the tensor
contraction of $T \times U$ is calculated:
\[
\sum_{i \ldots jkl \ldots m}{c^T_{i \ldots jk} c^U_{kl \ldots m} \mathbf{a}_i \otimes \ldots \otimes \mathbf{b}_j \otimes \mathbf{d}_l \otimes \ldots \otimes \mathbf{e}_m}
\]
where the resulting tensor is of rank equal to two less than the sum of the ranks of the input tensors;  the subtraction reflects the elimination of matching basis elements through summation during contraction.

For every curried multilinear map $g: A \to \ldots \to Y \to Z$, there is a tensor $T^g \in Z \otimes Y \otimes \ldots \otimes A$ encoding it \citep{bourbaki1989commutative,Lee:1997}. The application of a curried $n$-ary map $h : V_1 \to \ldots \to V_n \to W$ to input vectors $\mathbf{v_1} \in V_1,\,\ldots,\, \mathbf{v_n} \in V_n$ to produce output vector $\mathbf{w} \in W$ corresponds to the tensor contraction of the tensor $T^h \in W \otimes V_n \otimes \ldots \otimes V_1$ with the argument vectors:
\[
h(\mathbf{v_1})\ldots(\mathbf{v_n}) = T^h \times \mathbf{v_1} \times \ldots \times \mathbf{v_n}
\]
Using this correspondence between $n$-ary maps and tensors of rank
$n+1$
we can turn any
formal semantic model into a compositional distributional 
model. This is done by first running a type  inference algorithm on the generative rules and obtaining types, then assigning to each basic type a vector space and to each function type  a tensor space, and representing arguments by vectors and functions by
tensors,  finally, model function application  by tensor contraction.

To give an example, in the simple formal semantic model given above, a
type inference algorithm would provide us with basic types  $\semantics{N}$ and
$\semantics{S}$; we  assign vector
 spaces $N$ and $S$ to these respectively. Nouns and noun phrases are 
vectors in $N$, whereas sentences are vectors in $S$. Verb phrases map noun phrase interpretations to sentence interpretations, hence they are of type $\semantics{VP} : type(\semantics{NP}) \to type(\semantics{S})$,  in vector space terms we have 
$\semantics{VP} : N \to S$. Intransitive verbs map noun phrases to verb phrases, therefore have    the tensor form $T^{\textrm{vi}} \in S \otimes N$. Transitive verbs have type $\semantics{Vt} : \semantics{NP} \to
\semantics{VP}$, expanded to $\semantics{Vt} : N \to N
\to S$, giving us  the tensor form
$T^{\textrm{vt}} \in S \otimes N \otimes N$. Finally, adjectives are of type $\semantics{ADJ}: \semantics{N} \to \semantics{N}$, and hence have  the tensor form $T^{\textrm{adj}} \in N \otimes
N$. Putting all this together with tensor contraction ($\times$) as
function application, the vector meaning of our sample sentence ``angry
dogs chase furry cats''  is obtained by calculating the following operations, for lexical semantic vectors $T^{\textrm{cats}}$ and $T^{\textrm{dogs}}$, square matrices $T^{\textrm{furry}}$ and $T^{\textrm{angry}}$ , and a rank 3 tensor $T^{\textrm{chase}}$:

\[\left(T^{\textrm{chase}} \times
  \left(T^{\textrm{furry}} \times T^{\textrm{cats}} \right) \right)
\times \left( T^{\textrm{angry}} \times T^{\textrm{dogs}} \right)
\]
  
An important feature of the proposed approach is that elements
with the same syntactic category will always be represented by tensors
of the same rank and dimensionality. For examples, all phrases of type $S$ (namely sentences) will be represented by vectors with the same number of dimensions, making a
direct comparison of sentences with arbitrary syntactic structures
possible.

\section{Learning functions by multi-step regression}
\label{sec:learning}

The framework described above grants us the ability to determine the
rank of the tensors needed to encode functions, as well as their
dimensions relative to those of the vectors used to represent
arguments. It leaves open the question of how to learn tensors of
specific ranks. This, very much like in the case of the DisCoCat
framework of~\newcite{Coecke:etal:2010} from which it originated, is
intentional: There may be more than one suitable semantic
representation for arguments, functions, and sentences, and it is a
desirable feature that we may alternate between such representations
or combine them while leaving the mechanics of function composition
intact. Furthermore, there may be more than one way of learning the
tensors and vectors of a particular representation. Previous work on
learning tensors has been described independently by
\shortcite{Grefenstette:Sadrzadeh:2011a,Grefenstette:Sadrzadeh:2011b}
for transitive verbs, and by~\newcite{Baroni:Zamparelli:2010} for
adjective-noun constructions. In this section, we describe a new way
to learn such tensors, based on ideas from both aforementioned
approaches, namely that of multi-step regression.

Multi-step regression learning is a generalisation of linear
regression learning for tensors of rank 3 or higher, as procedures
already exist for tensors of rank 1 (lexical semantic vectors) and
rank 2 \citep{Baroni:Zamparelli:2010}. For rank 1 tensors, we suggest
learning vectors using any standard lexical semantic vector learning
model, and present sample parameters in Section
\ref{sec:semantic-space-construction} below. Learning rank 2 tensors
(matrices) can be treated as a multivariate multiple regression
problem, where the matrix components are chosen to optimise (in a
least square error sense) the mapping from training instances of input
(argument) to output (composed expression) vectors. Consider for
example the task of estimating the components of the matrix
representing an intransitive verb, that maps subject vectors to
(subject-verb) sentence vectors (Baroni and Zamparelli discuss the
analogous adjective-noun composition case):
\[ \mathbf{s} = V \times \mathbf{subj} \]
The weights of the matrix are estimated by least-squares regression
from example pairs of input subject and output sentence vectors
directly extracted from the corpus. For example, the matrix for
\emph{sing} is estimated from corpus-extracted vectors representing
pairs such as $<$\emph{mom}, \emph{mom sings}$>$, $<$\emph{child},
\emph{child sings}$>$, etc. Note that if the input and output vectors
are $n$ dimensional, we must estimate an $n \times{} n$ matrix, each
row corresponding to a separate regression problem (the $i$-th row
vector of the estimated matrix will provide the weights to linearly
combine the input vector components to predict the $i$-th output
vector component).  Regression is a supervised technique requiring
training data. However, we can extract the training data automatically
from the corpus and so this approach does not incur an extra knowledge
cost with respect to unsupervised methods.

\begin{figure*}[htb]
    \centering
    \includegraphics[scale=0.35]{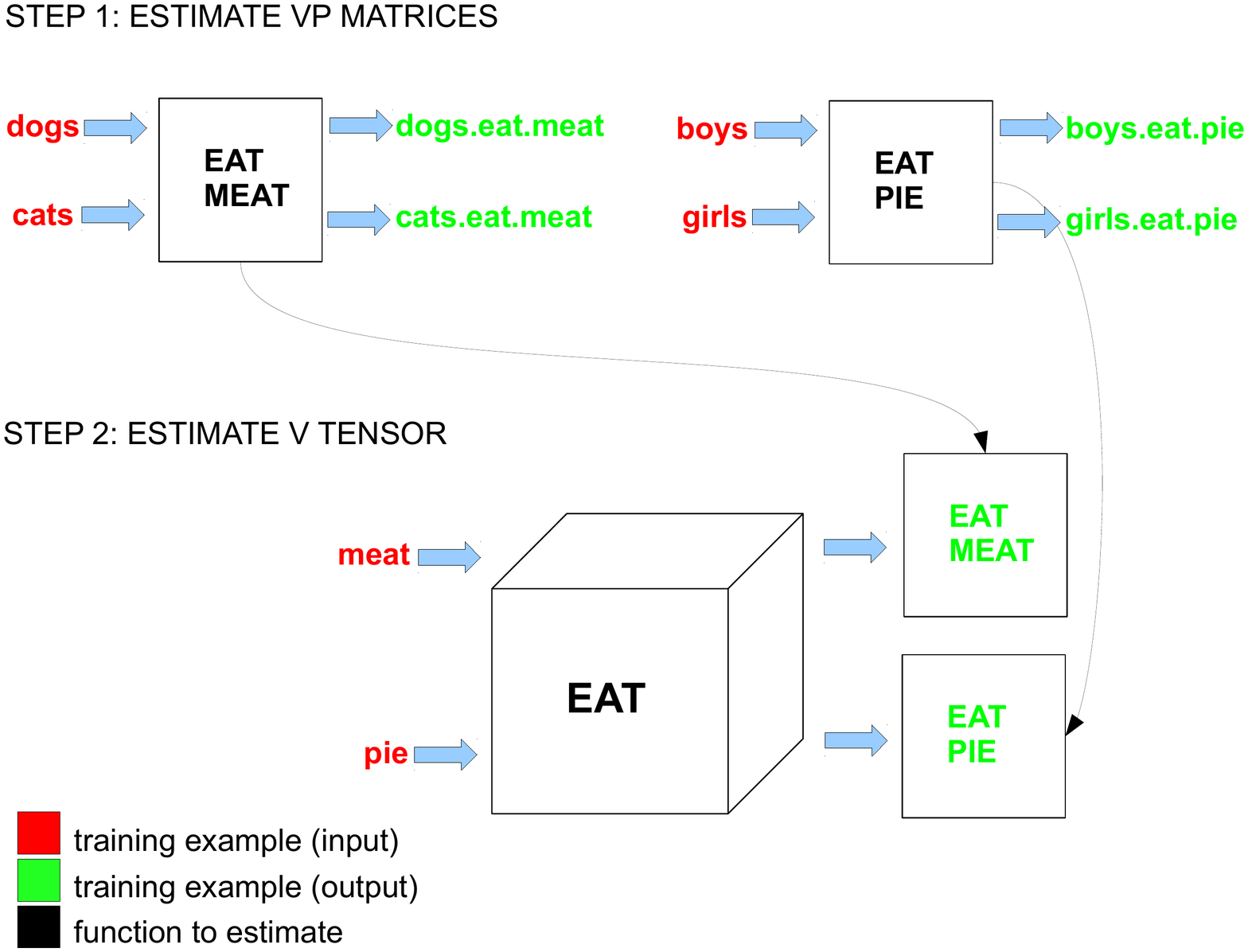}
    \caption{Estimating a tensor for \emph{eat} in two steps. We first
      estimate matrices for the VPs \emph{eat-meat}, \emph{eat-pie}
      etc.~by linear regression on input subject and output sentence
      vector pairs. We then estimate the tensor for \emph{eat} by
      linear regression with the matrices estimated in the previous
      step as output examples, and the vectors for the corresponding
      objects as input examples.}
    \label{fig:two-step-estimation}
\end{figure*}

Learning tensors of higher rank by linear regression involves
iterative application of the linear regression learning method
described above. The idea is to progressively learn the functions of
arity two or higher encoded by such tensors by recursively learning
the partial application of these functions, thereby reducing the
problem to the same matrix-learning problem as addressed by Baroni and
Zamparelli. To start with an example: the matrix-by-vector operation
of \newcite{Baroni:Zamparelli:2010} is a special case of the general
tensor-based function application model we are proposing, where a
`mono-argumental' function (intransitive verbs) corresponds to a rank
2 tensor (a matrix). The approach is naturally extended to
bi-argumental functions, such as transitive verbs, where the verb will
be a rank 3 tensor to be multiplied first by the object vector and
then by the subject, to return a sentence-representing vector:
\[ 
\mathbf{s} = V \times \mathbf{obj} \times \mathbf{subj} 
\]
The first multiplication of a $n\times{}n\times{}n$ tensor by a
$n$-dimensional vector will return a $n$-by-$n$ matrix (equivalent to
an intransitive verb, as it should be: both \emph{sings} and
\emph{eats meat} are VPs requiring a subject to be saturated). Note
that given $n$-dimensional input vectors, the $ij$-th $n$-dimensional
vector in the estimated tensor provides the weights to linearly
combine the input object vector components to predict the $ij$-th
output component of the unsaturated verb-object matrix. The matrix is
then multiplied by the subject vector to obtain a $n$-dimensional
vector representing the sentence. Again, we estimate the tensor
components by linear regression on input-output examples. In the first
stage, we apply linear regression to obtain examples of semi-saturated
matrices representing \emph{verb-object} constructions with a specific
verb. These matrices are estimated, like in the intransitive case,
from corpus-extracted examples of $<$subject, subject-verb-object$>$
pairs. After estimating a suitable number of such matrices for a
variety of objects of the same verb, we use pairs of corpus-derived
object vectors and the corresponding estimated verb-object matrices as
input-output pairs for another regression, where we estimate the verb
tensor components. The estimation procedure is schematically
illustrated for \emph{eat} in Fig.~\ref{fig:two-step-estimation}.

We can generalise this learning procedure to functions of arbitrary arity. Consider an $n$-ary function $f: X_1 \to \ldots \to X_n \to Y$. Let $L_i$ be the set of $i$-tuples $\{w^j_1,\ldots,w^j_i\}_{i\in[1,k]}$, where $k = |L_i|$, corresponding to the words which saturate the first $i$ arguments of $f$ in a corpus. For each tuple in some set $L_i$, let $f\ w^j_1\ \ldots\ w^j_i = f^j_i: X_{i+1} \to \ldots \to X_n \to Y$. Trivially, there is only one such $f^j_0$---namely $f$ itself---since $L_0 = \emptyset$ (as there are no arguments of $f$ to saturate for $i = 0$). The idea behind multi-step regression is to learn, at each step, the tensors for functions $f^j_i$ by linear regression over the set of pairs $(w^{j'}_{i+1},f^{j'}_{i+1})$, where the tensors $f^{j'}_{i+1}$ are the expected outcomes of applying $f^j_i$ to $w^{j'}_{i+1}$ and are learned during the previous step. We bootstrap this algorithm by learning the vectors in $Y$ of the set $\{f_n^j\}_j$ by treating the word which each $f^j_n$ models combined with the words of its associated tuple in $L_n$ as a single token. We then learn the vector for this token from the corpus using our preferred distributional semantics method. By recursively learning the sets of functions from $i=n$ down to $0$, we obtain smaller and smaller sets of increasingly de-saturated versions of $f$, which finally allow us to learn $f_0 =f$.

To specify how the set of pairs used for recursion is determined, let
there exist a function $super$ which takes the index of a tuple from
$L_i$ and returns the set of indices from $L_{i+1}$ which denote
tuples identical to the first tuple, excluding the last element:

\vspace{-.7cm}
\[
super: \mathbb{N} \times \mathbb{N} \to \mathcal{P}(\mathbb{N}) :: (i,j) \mapsto \{j' | \forall j' \in [1,k'].[w_1^j = w_1^{j'} \land \ldots \land w_i^j = w_i^{j'}]\}\ \text{where}\ k' = |L_{i+1}|
\]

\vspace{-.35cm}
\noindent
Using this function, the regression set for some $f_i^j$ can be defined as $\{(w^{j'}_{i+1},f^{j'}_{i+1})|j' = super(i,j)\}$.


While we just demonstrated how our model
generalises to functions of arbitrary arity, it remains to be seen if
in actual linguistic modeling there is an effective need for anything
beyond tri-argumental functions (ditransitive verbs).

\section{Experimental procedure}
\label{sec:experiments}

\subsection{Construction of distributional semantic vectors}
\label{sec:semantic-space-construction}

We extract co-occurrence data from the concatenation of the
Web-derived ukWaC corpus (\url{http://wacky.sslmit.unibo.it/}), a
mid-2009 dump of the English Wikipedia (\url{http://en.wikipedia.org})
and the British National Corpus
(\url{http://www.natcorp.ox.ac.uk/}). The corpus has been tokenised,
POS-tagged and lemmatised with TreeTagger
(\url{http://www.ims.uni-stuttgart.de/projekte/corplex/TreeTagger/})
and dependency-parsed with MaltParser
(\url{http://www.maltparser.org/}). It contains about 2.8 billion
tokens.

%
%
%
%
%
%

We collect vector representations for the top 8K most frequent nouns
and 4K verbs in the corpus, as well as for the subject-verb (320K) and
subject-verb-object (1.36M) phrases containing one of the verbs to be
used in one of the experiments below and subjects and objects from the
list of top 8K nouns. For all target items, we collect within-sentence
co-occurrences with the top 10K most frequent content words (nouns,
verbs, adjectives and adverbs), save for a stop list of the 300 most
frequent words. We extract co-occurrence statistics at the lemma
level, ignoring inflectional information. Following standard practice,
raw co-occurrence counts are transformed into statistically weighted
scores.  We tested various weighting schemes of the semantic space on
a word similarity task, observing
that 
non-negative pointwise mutual information (PMI) and local mutual
information (raw frequency count multiplied by PMI score) generally
outperform other weighting schemes by a large margin, and that PMI in
particular works best when combined with dimensionality reduction by
non-negative matrix factorization (described below). Consequently, we
pick PMI weighting for our experiments.

Reducing co-occurrence vectors to lower dimensionality is a common
step in the construction of distributional semantic models. Extensive
evidence suggests that dimensionality reduction does not affect, and
might even improve the quality of lexical semantic vectors
\citep{Bullinaria:Levy:2012,Landauer:Dumais:1997,Sahlgren:2006,Schuetze:1997}. In
our setting, dimensionality reduction is virtually necessary, since
working with 10K-dimensional vectors is problematic for the Regression
approach (see Section \ref{sec:composition-methods} below), that
requires learning matrices and tensors with dimensionalities which are
quadratic and cubic in the dimensionality of the input vectors,
respectively. We consider two dimensionality reduction methods, the
Singular Value Decomposition (SVD) and Non-negative Matrix
Factorization (NMF). SVD is the most common technique in
distributional semantics, and it was used by
\newcite{Baroni:Zamparelli:2010}. NMF is a less commonly adopted
method, but it has also been shown to be an effective dimensionality
reduction technique for distributional semantics
\citep{Dinu:Lapata:2010}. It has a fundamental advantage from our
point of view: The Multiply and Kronecker composition approaches (see
Section \ref{sec:composition-methods} below), because of their
multiplicative nature, cannot be meaningfully applied to vectors
containing negative values. NMF, unlike SVD, produces non-negative
vectors, and thus allows a fair comparison of all composition methods
in the same reduced space.\footnote{We ran the experiments reported
  below in full space for those models for which it was possible,
  finding that Multiply obtained better results there (approaching
  those of reduced-spaced Regression). This suggests that, although in
  our preliminary word similarity tests the original 10K-dimensional
  space and the two reduced spaces produced very similar results, it
  is still necessary to look for better low-dimensionality
  approximations of the full space.}

We perform the Singular Value Decomposition of the input matrix $X$:
$X = U\Sigma V^{T}$ and, like Baroni and Zamparelli and many others,
pick the first $k=300$ columns of $U\Sigma$ to obtain reduced
representations. Non-negative Matrix Factorization factorizes a $(m
\times n)$ non-negative matrix $X$ into two $(m \times k)$ and $(k
\times n)$ non-negative matrices: $X\approx WH$ (we normalize the
input matrix to $\sum_{i,j}X_{ij} = 1$ before applying NMF). We use
the Matlab implementation\footnote{Available at
  \url{http://www.csie.ntu.edu.tw/~cjlin/nmf/}.} of the projected
gradient algorithm proposed in \newcite{DBLP:journals/neco/Lin07},
which minimizes the squared error of Frobenius norm $F(W,H) = \|X -
WH\|^2_F$.  We set $k=300$ and we use $W$ as reduced representation of
input matrix $X$.\footnote{For both SVD and NMF, the latent
  dimensions are computed using a ``core'' matrix containing nouns and
  verbs only, subsequently projecting phrase vectors onto the same
  space. In this way, the dimensions of the reduced space do not
  depend on the ad-hoc choice of phrases required by our experiments.}

\subsection{Composition methods}
\label{sec:composition-methods}

\textbf{Verb} is a baseline measuring the cosine between the verbs in
two sentences as a proxy for sentence similarity (e.g., similarity of
\emph{mom sings} and \emph{boy dances} is approximated by the cosine
of \emph{sing} and \emph{dance}).  

We adopt the widely used and generally successful multiplicative and
additive models of \newcite{Mitchell:Lapata:2010} and
others. Composition with the \textbf{Multiply} and \textbf{Add
}methods is achieved by, respectively, component-wise multiplying and
adding the vectors of the constituents of the sentence we want to
represent. Vectors are normalised before addition, as this has
consistently shown to improve Add performance in our earlier
experiments.

\newcite{Grefenstette:Sadrzadeh:2011b} proposed a specific
implementation of the general DisCoCat approach to compositional
distributional semantics \citep{Coecke:etal:2010} that we call
\textbf{Kronecker} here. Under this approach, a transitive sentence is
a matrix $S$ derived from:
\[ S = (\mathbf{v} \otimes \mathbf{v}) \odot (\mathbf{subj} \otimes
\mathbf{obj}) \] That is, if nouns and verbs live in a $n$-dimensional
space, a transitive sentence is a $n\times{}n$ matrix given by the
component-wise multiplication of two Kronecker products: that of the
verb vector with itself and that of the subject and object
vectors. Grefenstette and Sadrzadeh show that this method outperforms
other implementations of the same formalism and is the current state
of the art on the transitive sentence task of
\newcite{Grefenstette:Sadrzadeh:2011a} we also tackle below.  For
intransitive sentences, the same approach reduces to component-wise
multiplication of verb and subject vectors, that is, to the Multiply
method.

Composition of nouns and verbs under the proposed (multi-step)
\textbf{Regression} model is implemented using Ridge Regression (RR)
\citep{Hastie:etal:2009}.  RR, also known as $L_{2}$ regularized
regression, is a different approach from the Partial Least Square
Regression (PLSR) method that was used in previous related work
\citep{Baroni:Zamparelli:2010,Guevara:2010} to deal with the
multicollinearity problem.  When multicollinearity exists, the matrix
$X^{T}X$ ($X$ here is the input matrix after dimensionality reduction)
becomes nearly singular and the diagonal elements of $(X^{T}X)^{-1}$
become quite large, which makes the variance of weights too large.  In
RR, a positive constant $\lambda$ is added to the diagonal elements of
$X^{T}X$ to strengthen its non-singularity.  Compared with PLSR, RR
has a simpler solution for the learned weight matrix $B = (X^{T}X +
\lambda I)^{-1}X^{T}Y$ and produces competitive results at a faster
speed.  For each verb matrix or tensor to be learned, we tuned the
parameter $\lambda$ by generalized cross-validation
\citep{golub1979generalized}. The objective function used for tuning
minimizes least square error when predicting corpus-observed sentence
vectors or intermediate VP matrices (the data sets we evaluate the
models on are \emph{not} touched during tuning!).
Training examples are found by combining the 8K nouns we have vectors
for (see Section \ref{sec:semantic-space-construction} above) with any
verb in the evaluation sets (see Sections \ref{sec:exp-transitives}
and \ref{sec:exp-intransitives} below) into subject-verb-(object)
constructions, and extracting the corresponding vectors from the
corpus, where attested (vectors are normalised before feeding them to
the regression routine). We use only example vectors with at least 10
non-0 dimensions before dimensionality reduction, and we require at
least 3 training examples per regression. For the first experiment
(intransitives), these (untuned) constraints result in an average of
281 training examples per verb. In the second experiment, in the
verb-object matrix estimation phase, we estimate on average 324
distinct matrices per verb, with an average of 15 training examples
per matrix. In the verb tensor estimation phase we use all relevant
verb-object matrices as training examples.\footnote{All materials and
  code used in these experiments that are not already publicly
  available can be requested to the first author.}


\section{Experiment 1: Predicting similarity judgments on intransitive sentences}
\label{sec:exp-transitives}

We use the test set of \newcite{Mitchell:Lapata:2008}, consisting of
180 pairs of simple sentences made of a subject and an intransitive
verb. The stimuli were constructed so as to ensure that there would be
pairs where the sentences have high similarity (\emph{the fire glowed}
vs.~\emph{the fire burned}) and cases where the sentences are
dissimilar while having a comparable degree of lexical overlap (\emph{the
  face glowed} vs.~\emph{the face burned}). The sentence pairs were
rated for similarity by 49 subjects on a 1-7 scale. Following Mitchell
and Lapata, we evaluate each composition method by the Spearman
correlation of the cosines of the sentence pair vectors, as predicted
by the method, with the individual ratings produced by the subjects
for the corresponding sentence pairs.


The results in table
\ref{tab:results}\subref{tab:intransitive-results} show that the
Regression-based model achieves the best correlation when applied to
SVD space, confirming that the approach proposed by Baroni and
Zamparelli for adjective-noun constructions can be successfully
extended to subject-verb composition. The Regression model also
achieves good performance in NMF space, where it is comparable to
Multiply.  Multiply was found to be the best model by Mitchell and
Lapata, and we confirm their results here (recall that Multiply can
also be seen as the natural extension of Kronecker to the intransitive
setting). The correlations attained by Add and Verb are considerably
lower than those of the other methods.


\section{Experiment 2: Predicting similarity judgments on transitive
  sentences}
\label{sec:exp-intransitives}

We use the test set of \newcite{Grefenstette:Sadrzadeh:2011a}, which was
constructed with the same criteria that Mitchell and Lapata applied,
but here the sentences have a simple transitive structure. An example
of a high-similarity pair is \emph{table shows result} vs.~\emph{table
  expresses result}; whereas \emph{map shows location} vs.~\emph{map
  expresses location} is a low-similarity pair. Grefenstette and
Sadrzadeh had 25 subjects rating each sentence. Model evaluation
proceeds like in the intransitive case.\footnote{Kronecker produces
  matrix representations of transitive sentences, so technically the
  similarity measure used for this method is the Frobenius inner
  product of the normalised matrices, equivalent to unfolding the
  matrices into vectors and computing cosine similarity.}



As the results in table
\ref{tab:results}\subref{tab:transitive-results} show, the Regression
model performs very well again, better than any other methods in NMF
space, and with a further improvement when SVD is used, similarly to
the first experiment.  The Kronecker model is also competitive,
confirming the results of Grefenstette and Sadrzadeh's
experiments. Neither Add nor Verb achieve very good results, although
even for them the correlation with human ratings is significant.

\begin{table}[t]
\centering
{\small
\subtable[Intransitive Sentences]{
  \centering
  \begin{tabular}{l|r}
    \emph{method}&$\rho$\\
    \hline
    Humans&0.40\\
    \hline
    Multiply.nmf&0.19\\
    Regression.nmf&0.18 \\
    Add.nmf&0.13\\
    Verb.nmf&0.08\\
    \hline
    Regression.svd&0.23\\
    Add.svd&0.11\\
    Verb.svd&0.06\\
  \end{tabular}
  \label{tab:intransitive-results}
  $\quad\quad\quad$
}
\subtable[Transitive Sentences]{
\centering
  \begin{tabular}{l|rl}
    \emph{method}&$\rho$\\
    \hline
    Humans&0.62\\
    \hline
    Regression.nmf&0.29 \\
    Kronecker.nmf&0.25\\
    Multiply.nmf&0.23\\
    Add.nmf&0.07\\
    Verb.nmf&0.04\\
    \hline
    Regression.svd&0.32\\
    Add.svd&0.12\\
    Verb.svd&0.08\\
  \end{tabular}
\label{tab:transitive-results}
  $\quad\quad\quad$
  }
}

\caption{Spearman correlation of composition methods with human similarity intuitions on two sentence similarity data sets (all correlations significantly above chance).  Humans is inter-annotator correlation.  The multiplication-based Multiply and Kronecker methods are not well-suited for the SVD space (see Section \ref{sec:semantic-space-construction}) and their performance is reported in NMF space only. Kronecker is only defined for the transitive case, Multiply functioning also as its intransitive-case equivalent (see Section \ref{sec:composition-methods}).}
\label{tab:results}
\end{table}

\section{General discussion of the results}
\label{sec:discussion}


The results presented here show that our iterative linear regression
algorithm outperforms the leading multiplicative method on
intransitive sentence similarity when using SVD (and it is on par with
it when using NMF), and outperforms both the multiplicative method and
the leading Kronecker model in predicting transitive sentence
similarity. Additionally, the multiplicative model, while commendable
for its extreme simplicity, is of limited general interest, since it
cannot take word order into account. We can trivially make this model
fail by testing it on transitive sentences with subject and object
inverted: For Multiply, \emph{pandas eat bamboo} and \emph{bamboo eats
  pandas} are identical statements, whereas for humans they are
obviously very different.

Confirming what Grefenstette and Sadrzadeh found, we saw that
Kronecker performs very well also in our experimental setup (although
not as well as Regression). The main advantage of Kronecker over
Regression lies in its simplicity: there is no training involved, all
it takes is two outer vector products and a component-wise
multiplication. However, as pointed out
by~\newcite{Grefenstette:Sadrzadeh:2011b}, this method is ad hoc
compared to the linguistically motivated Categorical method they
initially presented in~\newcite{Grefenstette:Sadrzadeh:2011a}. It is
conceivable that the Kronecker model's good performance is primarily
tied to the nature of the evaluation data-set, where only verbs change
while subject and object stay the same in sentence pairs.

While our regression-based model's estimation procedure is
considerably more involved than for Kronecker, the model has much to
recommend it, both from a statistical and from a linguistic point of
view. On the statistical side, there are many aspects of the
estimation routine that could be tuned on automatically collected
training data, thus bringing up the Regression model performance. We
could for example harvest a larger number of training phrases (not
limiting them to those that contain nouns from the 8K most frequent in
the corpus, as we did), or \emph{vice versa} limit training to more
frequent phrases, whose vectors are presumably of better
quality. Moreover, Ridge Regression is only one of of many estimation
techniques that could be tried to come up with better matrix and
tensor weights. %
On the linguistic side, the model is clearly motivated as an
instantiation of the vector-space ``dual'' of classic composition by
function application via the tensor contraction operation, as
discussed in Section \ref{sec:model} above.  Moreover, Regression
produces vectors of the same dimensionality for sentences formed with
intransitive and transitive verbs, whereas for Kronecker, if the
former are $n$-dimensional vectors, the second are $n\times{}n$
matrices. Thus, under Kronecker composition, sentences with
intransitive and transitive verbs are not directly comparable, which
is counter-intuitive (being able to measure the similarity of, say,
\emph{kids sing} and \emph{kids sing songs} is both natural and
practically useful).

Finally, we remark that in both experiments SVD-reduced vectors lead
to Regression models outperforming their NMF counterparts. Regression,
unlike the multiplication-based models, is not limited to non-negative
vectors, and it can thus harness the benefits of SVD reduction
(although of course it is precisely because of the large regression
problems we must solve that we need to perform dimensionality
reduction at all!). 

\section{Conclusion}
\label{sec:conclusion}

The main advances introduced in this paper are as follows. First, we
discussed a tensor-based compositional distributional semantic
framework in the vein of that of~\newcite{Coecke:etal:2010} which has
the compositional mechanism of~\newcite{Baroni:Zamparelli:2010} as a
specific case, thereby uniting both lines of research in a common
framework. Second, we presented a generalisation of Baroni and
Zamparelli's matrix learning method to higher rank tensors, allowing
us to induce the semantic representation of functions modelled in this
framework. Finally, we evaluated this new semantic tensor learning
model against existing benchmark data-sets provided
by~\newcite{Mitchell:Lapata:2008}
and~\newcite{Grefenstette:Sadrzadeh:2011a}, and showed it to
outperform other models. We furthermore claim that the generality of
our extended regression method allows it to capture more information
than the multiplicative and Kronecker models, and will allow us to
canonically model more complex and subtle relations where argument
order and semantic roles matter more, such as quantification, logical
operations, and ditransitive verbs.

Among the plans for future work, we intend to improve regression-based
tensor estimation, focusing in particular on automated ways to choose
informative training examples. On the evaluation side, we want to
construct a larger test set to directly compare sentences with
different argument counts (e.g., transitive vs.~intransitive
constructions) and word orders (e.g., sentences with subject and
object inverted), as well as extending modeling and evaluation to
other syntactic structures and types of function application
(including the challenging cases we listed in the previous
paragraph). We want moreover to test the Regression model against the
Categorical model of~\newcite{Grefenstette:Sadrzadeh:2011a} and to
design evaluation scenarios allowing a direct comparison with the
MV-RNN model of~\newcite{socherEMNLP12}.

\section*{Acknowledgments}
Edward Grefenstette is supported by EPSRC Project \emph{A Unified Model of Compositional and Distributional Semantics: Theory and Applications} (EP/I03808X/1). Georgiana Dinu and Marco Baroni are partially supported by the ERC 2011 Starting Independent Research Grant to the COMPOSES project (n.~283554). Mehrnoosh Sadrzadeh is supported by an EPSRC Career Acceleration Fellowship (EP/J002607/1). 

\bibliographystyle{chicago}

\newpage

{\footnotesize
\bibliography{marco,ed,georgiana} }

\end{document}